\documentclass[11pt]{article}
\usepackage{acl}

\usepackage{times}
\usepackage{latexsym}

\usepackage[T1]{fontenc}

\usepackage[utf8]{inputenc}

\usepackage{microtype}

\usepackage{inconsolata}

\usepackage{graphicx}
\usepackage{amssymb}
\usepackage{amsmath}
\usepackage{booktabs}
\usepackage{subcaption}
\usepackage[table]{xcolor}
\definecolor{lightgray}{gray}{0.85}
\usepackage{multirow}
\usepackage{float}

 \usepackage[capitalize,noabbrev]{cleveref}
\crefname{section}{Sec.}{Secs.}
\Crefname{section}{Appendix}{Appendices}
\crefname{table}{Table}{Tables}
\crefname{figure}{Fig.}{Figs.}
\crefname{algocf}{alg.}{algs.}
\Crefname{algocf}{Algorithm}{Algorithms}

\title{Internalize the Temperature: On-Policy Self-Distillation as Policy Reheater for Reinforcement Learning}

\author{
Xuewei Yang$^{1 \ast}$,
Jiachen Yu$^{1}$\thanks{\, \, Equal contribution.},
Jie Wu$^{1}$,
Shaoning Sun$^1$,
Junjie Wang$^{1 \dag}$,
Yujiu Yang$^1$\thanks{\, \, Corresponding author.}
\\
$^1$Tsinghua University\\
\texttt{\{yangxw25,yu-jc21\}@mails.tsinghua.edu.cn}\\
\texttt{\{wangjunjie,yang.yujiu\}@sz.tsinghua.edu.cn}
}

\begin{document}
\maketitle
\begin{abstract}
Reinforcement learning from verifiable rewards improves the reasoning ability of large language models, but often suffers from \textit{entropy collapse}, in which increasingly concentrated policies reduce rollout diversity and useful learning signals.
Existing remedies either constrain the RL objective (e.g., entropy regularization) or adjust sampling temperature during rollout collection, but these interventions remain external to the model parameters.
We propose Temperature-Scaled On-Policy Self-Distillation (TS-OPSD), a lightweight policy reheating method that internalizes the exploratory effect of temperature into model parameters.
Starting from an entropy-collapsed RL checkpoint, TS-OPSD constructs a self-teacher by applying high-temperature scaling to the model’s own logits, then distills the resulting smoother distribution back into the student.
This policy reheating requires no external teacher, privileged data, or additional inference cost.
Experiments on Qwen3-4B-Base and Qwen3-8B-Base show that policy reheating yields a stronger initialization for continued RL than both standard continued RL and rollout-level temperature reheating.
Further analyses show that TS-OPSD mainly reduces output sharpness while preserving intermediate representations, top candidate sets, and reasoning capability. 
These results suggest that entropy restoration can serve as a simple post-collapse intervention for extending reasoning-oriented RL.
\end{abstract}

\section{Introduction}

\begin{figure}[t]
    \centering
    \includegraphics[width=\linewidth]{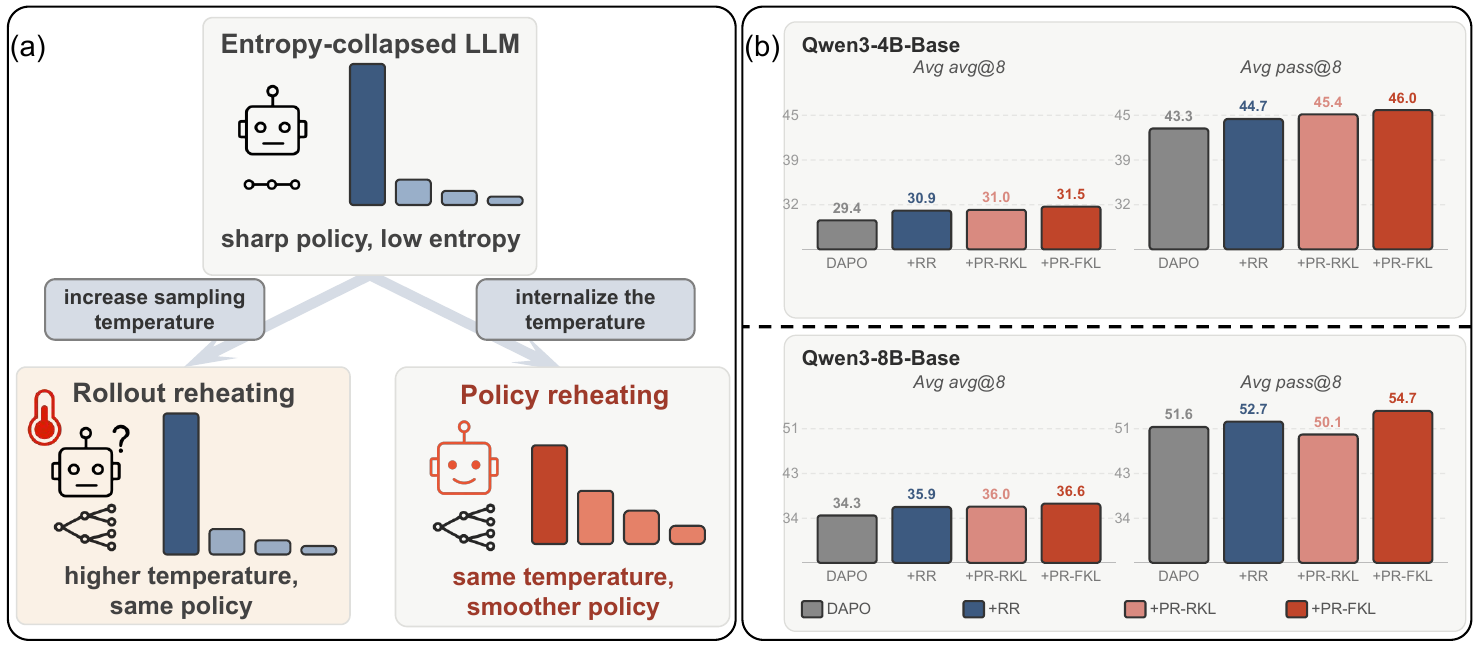}
    \caption{
    \textbf{(a)} Rollout reheating increases sampling temperature without modifying the policy itself, whereas policy reheating (TS-OPSD) internalizes temperature directly into the model parameters.
    \textbf{(b)} After applying policy reheating to an entropy-collapsed checkpoint and resuming RL, DAPO~w/~PR-FKL consistently outperforms both the DAPO baseline and rollout reheating (DAPO~w/~RR) on Qwen3-4B-Base and Qwen3-8B-Base across Avg avg@8 and Avg pass@8.
    Avg avg@8 and Avg pass@8 are macro-averages across MATH-500, AIME24, AIME25, AIME26, and OlymMATH.
    }
    \label{fig:teaser}
\end{figure}

Reinforcement learning from verifiable rewards (RLVR) has become a key approach for improving the reasoning abilities of large language models, especially in tasks involving mathematical and logical reasoning \citep{deepseekmath,deepseekr1,yu2025dapoopensourcellmreinforcement}.
However, this improvement often comes with a significant training bottleneck: as policy optimization progresses, the model distribution becomes increasingly concentrated, reducing rollout diversity and leaving little useful gradient information for further improvement, since effective learning signals are concentrated on a small fraction of high-entropy tokens at critical reasoning branch points \citep{cui2025entropymechanismreinforcementlearning,wang20258020rulehighentropyminority}.
Therefore, maintaining or restoring the ability to explore after entropy collapse is crucial for extending RL beyond its early stages.

Existing approaches mainly address this issue by intervening during RL optimization or rollouts.
One approach modifies the RL objective or update rule through entropy regularization, entropy-preserving updates, or adaptive clipping \citep{petrenko2026entropypreservingreinforcementlearning,xi2025bapostabilizingoffpolicyreinforcement,cheng2025reasoningexplorationentropyperspective}.
Another treats sampling temperature as a control parameter during rollout collection, either at the trajectory level or through learned token-level temperature policies \citep{dang2026temperature,zhou2026lookinwardexploreoutward}.
However, both classes of intervention remain external to the policy itself: objective-level methods impose additional constraints that trade off against reward maximization, whereas temperature-based rollout methods alter the behavior policy without changing the model's default distribution, resulting in a mismatch between the behavior policy and the optimized policy that may limit the effectiveness of the update.
This raises a natural question: can the exploratory effect of temperature be incorporated directly into the model parameters, so that the resulting policy serves as a better initialization for further reinforcement learning?

We propose \textbf{Temperature-Scaled On-Policy Self-Distillation (TS-OPSD)}, a lightweight method for restoring an entropy-degraded policy by distilling its temperature-scaled distribution back into its parameters.
Given the current model logits, we construct a self-teacher by applying a high-temperature softmax transformation, requiring no external model, special context, or additional inference steps.
We then optimize a full-logits KL objective from this self-teacher to the student, transferring the smoother, higher-entropy distribution into the model weights.
We refer to this process as \textit{internalizing temperature}: rather than using temperature as a temporary decoding control, TS-OPSD transforms its exploratory effect into a parameterized characteristic of the policy itself.
~\cref{fig:teaser} illustrates the distinction between rollout reheating and policy reheating, and summarizes the resulting performance gains on Qwen3-4B-Base and Qwen3-8B-Base.

We study TS-OPSD in the context of RL entropy collapse.
Starting from a collapsed RL checkpoint, TS-OPSD restores policy entropy while maintaining downstream reasoning accuracy.
When RL is continued from the restored checkpoint, the model achieves better performance than both standard continued RL and the method that merely increases sampling temperature during rollout collection.
This contrast suggests that the benefit of TS-OPSD does not stem merely from a noisier rollout process, but from the resulting smoother policy distribution, which serves as a better foundation for subsequent optimization.

Further analyses show that TS-OPSD performs support-preserving probability redistribution: it softens output distributions while leaving intermediate representations, top candidate sets, and downstream reasoning capability largely intact. 
We also characterize the training dynamics of TS-OPSD and analyze the effects of KL direction and temperature scaling on entropy recovery.

\section{Related Work}
\label{sec:related_work}

\begin{table*}[t]
\centering
\resizebox{\textwidth}{!}{
\begin{tabular}{lcccccccccccc}
\toprule
& \multicolumn{2}{c}{MATH-500} & \multicolumn{2}{c}{AIME24} & \multicolumn{2}{c}{AIME25} & \multicolumn{2}{c}{AIME26} & \multicolumn{2}{c}{OlymMATH} & \multicolumn{2}{c}{Avg} \\
\cmidrule(lr){2-3}\cmidrule(lr){4-5}\cmidrule(lr){6-7}\cmidrule(lr){8-9}\cmidrule(lr){10-11}\cmidrule(lr){12-13}
Model & avg@8 & pass@8 & avg@8 & pass@8 & avg@8 & pass@8 & avg@8 & pass@8 & avg@8 & pass@8 & avg@8 & pass@8\\
\midrule
\multicolumn{11}{l}{\textit{Qwen3-4B-Base}} \\
Base              & 62.38 & 89.00 & 8.75  & 36.67 & 7.92  & 23.33 & 6.25  & 16.67 & 2.94 & 14.00 & 17.65 & 35.93 \\
+DAPO~\citep{yu2025dapoopensourcellmreinforcement}              & 86.10 & 94.20 & 22.08 & 40.00 & 18.33 & 36.67 & 14.17 & 30.00 & 6.31 & 15.50 & 29.40 & 43.27 \\
\qquad w/ Entropy Reg       & 84.25 & \textbf{95.40} & 21.25 & 40.00 & 20.83 & 36.67 & \underline{16.67} & 26.67 & 6.69 & 21.00 & 29.94 & 43.95 \\
\qquad w/ Entropy Adv      & 85.40 & 94.20 & 23.33 & 36.67 & \underline{21.25} & 40.00 & \underline{16.67} & \textbf{33.33} & 6.81 & \underline{21.50} & 30.69 & 45.14 \\
\qquad w/ RR        & 85.98 & \underline{95.00} & \underline{24.17} & 40.00 & \textbf{22.50} & \textbf{40.00} & 15.00 & 30.00 & 6.63 & 18.50 & 30.86 & 44.70 \\
\qquad w/ Adaptive RR      & 85.80 & 94.40 & 22.08 & 40.00 & 17.92 & 36.67 & 15.83 & 30.00 & 7.00 & 19.50 & 29.73 & 44.11 \\
\qquad w/ PR-RKL (\textit{ours})    & \underline{86.20} & 94.20 & \textbf{25.00} & \textbf{43.33} & 20.83 & 36.67 & 15.83 & \textbf{33.33} & \underline{7.06} & 19.50 & \underline{30.98} & \underline{45.41} \\
\rowcolor{lightgray}
\qquad w/ PR-FKL (\textit{ours})    & \textbf{86.48} & 94.40 & \underline{24.17} & \textbf{43.33} & \underline{21.25} & \textbf{40.00} & \textbf{17.08} & 30.00 & \textbf{8.31} & \textbf{22.50} & \textbf{31.46} & \textbf{46.05} \\
\midrule
\multicolumn{11}{l}{\textit{Qwen3-8B-Base}} \\
Base              & 60.93 & 89.80 & 9.58  & 26.67 & 8.75  & 26.67 & 6.67  & 20.00 & 3.38 & 12.00 & 17.86 & 35.03 \\
+DAPO~\citep{yu2025dapoopensourcellmreinforcement}              & 87.65 & 96.00 & 24.58 & \underline{50.00} & 23.33 & \textbf{46.67} & 25.42 & \underline{43.33} & 10.25 & 22.00 & 34.27 & 51.60 \\
\qquad w/ Entropy Reg       & 88.55 & 95.80 & \underline{28.75} & \underline{50.00} & 25.00 & 36.67 & 24.17 & 40.00 & 10.75 & 26.50 & 35.44 & 49.79 \\
\qquad w/ Entropy Adv       & 88.78 & \underline{96.20} & \underline{28.75} & 46.67 & 25.42 & 40.00 & 19.58 & 40.00 & 9.00 & 21.50 & 34.31 & 48.87 \\
\qquad w/ RR        & \underline{89.33} & \textbf{96.80} & 25.42 & \underline{50.00} & 25.83 & \textbf{46.67} & \textbf{27.50} & \underline{43.33} & \underline{11.56} & 26.50 & 35.93 & \underline{52.66} \\
\qquad w/ Adaptive RR       & 88.53 & 95.80 & 24.58 & 43.33 & 25.42 & 43.33 & 23.75 & 40.00 & 11.25 & \underline{27.00} & 34.71 & 49.89 \\
\qquad w/ PR-RKL (\textit{ours})   & 88.58 & 95.60 & 28.33 & \underline{50.00} & \underline{26.25} & 40.00 & \underline{25.83} & 40.00 & 11.13 & 25.00 & \underline{36.02} & 50.12 \\
\rowcolor{lightgray}
\qquad w/ PR-FKL (\textit{ours})    & \textbf{89.38} & \underline{96.20} & \textbf{29.17} & \textbf{53.33} & \textbf{27.50} & \textbf{46.67} & 25.00 & \textbf{50.00} & \textbf{11.93} & \textbf{27.50} & \textbf{36.60} & \textbf{54.74} \\
\bottomrule
\end{tabular}
}
\caption{
Main results on MATH-500, AIME24, AIME25, AIME26 and OlymMATH.
avg@8 and pass@8 are reported for each benchmark and their macro-average (Avg).
\textbf{Bold} indicates the best result; \underline{underline} indicates the second best.
}
\label{tab:main}
\end{table*}

\textbf{Entropy in RLVR.}
Reinforcement learning has become a dominant post-training paradigm for LLM reasoning, yet entropy collapse, where the policy distribution sharpens as training proceeds, reducing rollout diversity and starving the optimizer of gradient signal, has emerged as a recurring bottleneck.
DAPO introduces dynamic sampling to sustain exploration \citep{yu2025dapoopensourcellmreinforcement}; \citet{cui2025entropymechanismreinforcementlearning} traces its mechanistic connections to reward improvement and training saturation; and \citet{wang20258020rulehighentropyminority} shows that effective learning signal concentrates on a small fraction of high-entropy tokens at critical reasoning branch points.
Remediation efforts include entropy-preserving updates, adaptive clipping, and entropy regularization \citep{petrenko2026entropypreservingreinforcementlearning,xi2025bapostabilizingoffpolicyreinforcement,cheng2025reasoningexplorationentropyperspective}.

\textbf{On-Policy Distillation and Self-Distillation.}
Knowledge distillation has shifted from static offline regimes toward on-policy settings, where GKD supervises students on their own generated trajectories \citep{agarwal2024onpolicydistillationlanguagemodels} and MiniLLM optimizes sequence-level reverse KL for stable generative distillation \citep{gu2026minillmonpolicydistillationlarge}; recent work further analyzes failure modes, including support-set approximation, sampled-token bias, and KL direction \citep{fu2026revisitingonpolicydistillationempirical,li2026rethinkingonpolicydistillationlarge}.
On-policy self-distillation (OPSD) removes the need for an external teacher by having the same model play both roles under different conditions: Self-Distilled Reasoner uses privileged-information conditioning \citep{zhao2026selfdistilledreasoneronpolicyselfdistillation}, while related variants exploit context, concise-reasoning prompts, or reward signals as the conditioning variable \citep{ye2026onpolicycontextdistillationlanguage,sang2026crispcompressedreasoningiterative,hubotter2026reinforcementlearningselfdistillation,yang2026selfdistilledrlvr}.
\citet{jin2026entropyawareonpolicydistillationlanguage} further shows that forward KL better preserves high-entropy teacher support than reverse KL.

\textbf{Temperature in Knowledge Distillation and Model Reasoning.}
Temperature has long served as a core mechanism in knowledge distillation, where high-temperature softmax targets expose dark knowledge beyond hard labels \citep{hinton2015distillingknowledgeneuralnetwork}; subsequent work shows it can be scheduled or adapted rather than fixed \citep{jafari2021annealingknowledgedistillation,li2022curriculumtemperatureknowledgedistillation,wei2024dynamictemperatureknowledgedistillation}, with its effect depending on teacher quality and training configuration \citep{frank2026unifiedrevisittemperatureclassificationbased}.
At inference time, \citet{zhang2024edtimprovinglargelanguage} adjusts decoding temperature dynamically according to model uncertainty, and \citet{wu2025roletemperaturesamplingtesttime} shows that different sampling temperatures unlock competence on different reasoning subsets, expanding the effective reasoning frontier.
In reinforcement learning, temperature has also been treated as a learnable or adaptive exploration-control mechanism during rollout collection \citep{dang2026temperature,zhou2026lookinwardexploreoutward}.
TS-OPSD differs from these approaches in that it internalizes the exploratory effect of temperature directly into model parameters, rather than using it as a transient sampling control.

\section{Method}
\subsection{Preliminaries}

\paragraph{Temperature scaling.}
Temperature scaling is a standard technique for controlling the sharpness of a language model's output distribution at inference time.
Given logits $z_\theta \in \mathbb{R}^{|V|}$, the temperature-scaled distribution is defined as
\begin{equation}
\label{eq:temp-scale}
q_T(v \mid x_{<t}) = \frac{\exp(z_{\theta,v}/T)}{\sum_{u}\exp(z_{\theta,u}/T)},
\end{equation}
where $T$ is the temperature parameter. 
Higher temperatures produce flatter, more exploratory distributions, while lower temperatures sharpen the distribution toward the model's top predictions.

\paragraph{KL divergence in LLM distillation.}
KL divergence is a standard measure of discrepancy between two distributions and has long been used as the alignment objective in language-model knowledge distillation~\citep{kl-divergence}.
Given a teacher distribution $p(\cdot \mid x_{<t})$ and a student distribution $q_\theta(\cdot \mid x_{<t})$, the two most common forms are forward KL (FKL) and reverse KL (RKL):
\begin{equation}
\small
D_{\text{FKL}}(p \| q_\theta)
= \sum_{v} p(v \mid x_{<t}) \log \frac{p(v \mid x_{<t})}{q_\theta(v \mid x_{<t})},
\end{equation}

\begin{equation}
\small
D_{\text{RKL}}(q_\theta \| p)
= \sum_{v} q_\theta(v \mid x_{<t}) \log \frac{q_\theta(v \mid x_{<t})}{p(v \mid x_{<t})}.
\end{equation}
The choice between FKL and RKL has been extensively discussed in the context of LLM distillation, with no consensus yet reached on which is universally preferable \citep{gu2026minillmonpolicydistillationlarge,agarwal2024onpolicydistillationlanguagemodels,wen-etal-2023-f,wu-etal-2025-rethinking}.

\subsection{Temperature-Scaled On-Policy Self-Distillation}

Unlike existing OPSD variants that construct the teacher distribution through external context, privileged information, or reward conditioning \citep{zhao2026selfdistilledreasoneronpolicyselfdistillation,ye2026onpolicycontextdistillationlanguage,hubotter2026reinforcementlearningselfdistillation,yang2026selfdistilledrlvr}, TS-OPSD derives the teacher entirely from the student's own logits via temperature scaling.
At each training step, given the current student logits $z_\theta$, we construct the self-teacher $p_T$ as in \cref{eq:temp-scale} with stop-gradient teacher logits.
The student is then trained to minimize the KL divergence between this high-temperature self-teacher and its own current distribution, using on-policy rollouts sampled from $q_\theta$ itself.
No additional model, inference pass, or external supervision is required.

A key structural property of this construction is that $p_T$ is a monotone power transform of $q_\theta$: since $p_T(v) \propto q_\theta(v)^{1/T}$, the token ranking is fully preserved between teacher and student, and the two distributions share the same local support.
The teacher differs from the student only in distributional sharpness, not in semantic preference ordering.
This rank-preserving coupling is structurally distinct from standard KD, where arbitrary teacher-student mismatch can cause tail tokens to dominate the gradient.
As we discuss in \cref{sec:kl_analysis}, this coupling also narrows the behavioral gap between FKL and RKL compared to standard distillation settings.
By training the student to match its own high-temperature distribution, TS-OPSD transfers the smoother, higher-entropy output behavior into the model parameters without modifying the decoding procedure or introducing any train-inference mismatch---we refer to this process as \textit{internalizing the temperature}.

TS-OPSD is designed to serve as a lightweight post-collapse intervention between RL stages:
\begin{equation}
\theta_{\text{RL-collapse}}
\xrightarrow{\text{high-}T\text{ OPSD}}
\theta_{\text{reheat}}
\xrightarrow{\text{continued RL}}
\theta_{\text{final}}.
\end{equation}
Starting from a collapsed checkpoint, TS-OPSD restores policy entropy while preserving the reasoning preferences already acquired during training, providing a smoother initialization for continued RL.
Unlike entropy regularization or temperature-based rollout methods, TS-OPSD directly modifies the policy distribution in a knowledge-preserving manner, decoupling entropy restoration from the RL objective and eliminating train-inference inconsistency.

\section{Experiments}

\begin{figure}[!t]
    \centering
    \includegraphics[width=\linewidth]{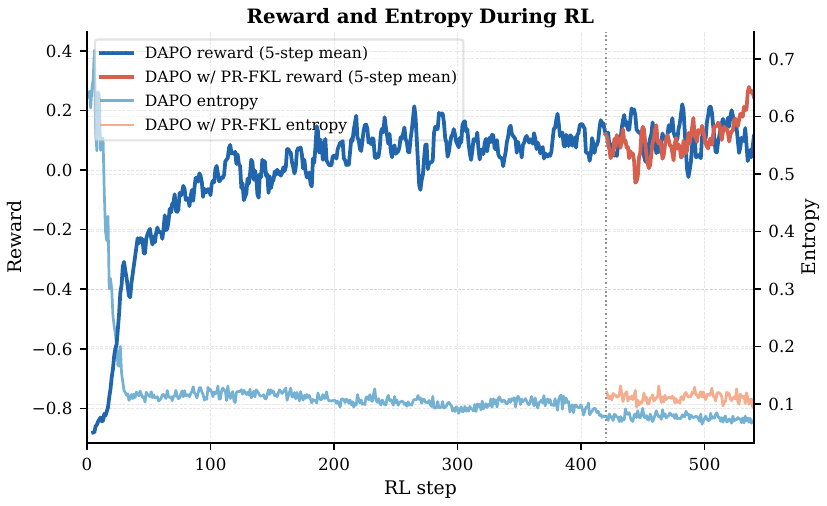}
    \caption{
        Reward and mean token entropy during RL training on Qwen3-8B-Base.
        The dotted vertical line marks the point at which policy reheating (PR-FKL) is applied to the collapsed checkpoint.
        After reheating, DAPO w/ PR-FKL (red) resumes from a higher-entropy initialization and finally achieves a higher reward than standard DAPO (blue) in the continued training phase.
    }
    \label{fig:reward_entropy}
\end{figure}

\subsection{Experimental Setup}
\label{sec:exp_setup}

\textbf{Models and training data.}
We conduct experiments on Qwen3-4B-Base and Qwen3-8B-Base.
All models are trained on DAPO-Math-17K \citep{yu2025dapoopensourcellmreinforcement} using the DAPO training recipe as the base RL procedure, implemented on top of verl \citep{sheng2024hybridflow}.

\textbf{Evaluation.}
We evaluate on MATH-500 \citep{hendrycksmath2021}, AIME24, AIME25, AIME26 and OlymMATH \citep{sun2026challengingboundariesreasoningolympiadlevel}(English easy and hard splits combined).
For each benchmark, we report avg@8 (mean accuracy over 8 samples per prompt) and pass@8 (probability of at least one correct answer).
The Avg column reports the macro-average across all five benchmarks.
All evaluations use sampling temperature $T=1.0$ and top-$p=0.7$.

\textbf{Baselines.}
We compare the following settings:
Base is the pretrained checkpoint without RL training.
DAPO is standard RL trained until entropy collapse, serving as the primary baseline.
DAPO w/ Entropy Reg adds an entropy bonus to the DAPO objective with coefficient 0.001.
DAPO w/ Entropy Adv uses an entropy-aware advantage with $\alpha=0.4$ and $\kappa=2$, following \citet{zhou2026lookinwardexploreoutward}.
DAPO w/ Adaptive RR dynamically schedules rollout temperature following the adaptive-temperature strategy of \citet{dang2026temperature}.
DAPO w/ RR (Rollout Reheating) resumes RL from the collapsed checkpoint with rollout sampling temperature raised to $T=1.2$, keeping all other hyperparameters unchanged. 
The temperature is chosen to match the teacher's temperature used in policy reheating, enabling a direct comparison between modifying the sampling process and modifying the policy parameters.
DAPO w/ PR-FKL and DAPO w/ PR-RKL (Policy Reheating, our method) apply high-temperature OPSD with forward KL and reverse KL, respectively, at the collapsed checkpoint, then resume standard RL.

\textbf{OPSD training details.}
High-temperature OPSD uses teacher temperature $T=1.2$, learning rate $1\times10^{-6}$, effective batch size 1024, and full-vocabulary KL as the training objective, implemented via the GKD framework in swift \citep{zhao2025swiftascalablelightweightinfrastructure}.
Rollouts are sampled on-policy from the student at each step with no additional regularization.
The reheating intervention is applied after the initial entropy decline has stabilized and the policy has entered a persistently low-entropy regime; in our setting, the selected checkpoint has mean token entropy below 0.08 nats.
This value is a setting-specific operational choice rather than a universal collapse threshold.
During reheating, we select a checkpoint in the stable phase after the characteristic entropy jump (\cref{sec:kl_analysis}), while task accuracy remains stable and before entropy begins to accelerate under prolonged OPSD (\cref{app:long_horizon_opsd}).
In our experiments this corresponds to recovery above 0.12 nats.
Full hyperparameters for RL training, OPSD, and evaluation are provided in ~\cref{app:hyperparams}.

\subsection{Main Results}
\label{sec:main_results}

~\cref{tab:main} reports results for Qwen3-4B-Base and Qwen3-8B-Base across all evaluation benchmarks; ~\cref{fig:reward_entropy} shows the corresponding reward and entropy dynamics during RL training on the 8B model.

\paragraph{Entropy collapse stalls RL.}
As shown in ~\cref{fig:reward_entropy}, the entropy of the DAPO baseline drops sharply in the early training phase and remains suppressed throughout, with the rate of reward improvement slowing considerably before training ends.
This is consistent with the structural bottleneck described in ~\cref{sec:related_work}: once the policy concentrates on a narrow set of trajectories, gradient signals diminish and further optimization becomes ineffective.

\paragraph{Reheating variants improve over RL.}
After the reheating intervention, DAPO w/ PR-FKL resumes from a higher-entropy initialization and achieves a higher reward than standard DAPO in the continued training phase (\cref{fig:reward_entropy}).
Rollout reheating still outperforms standard DAPO across nearly all benchmarks and model sizes (\cref{tab:main}), with gains most consistent on the harder benchmarks, confirming that restoring exploratory capacity after entropy collapse enables the policy to reach higher performance ceilings.
Among the two KL objectives, PR-FKL consistently outperforms PR-RKL; we analyze this difference in ~\cref{sec:kl_analysis}.

\paragraph{Policy reheating outperforms rollout reheating and entropy-preserving methods.}
Compared with standard DAPO, the entropy-preserving baselines yield only marginal and inconsistent improvements across model scales and evaluation metrics.
DAPO w/ RR already provides non-trivial gains over DAPO, particularly on pass@8 metrics, consistent with prior work on adaptive rollout temperature \citep{dang2026temperature,zhou2026lookinwardexploreoutward}.
DAPO w/ PR-FKL further outperforms DAPO w/ RR across most benchmarks, with the advantage most pronounced on the 8B model.
This gap suggests that the benefit of policy reheating is not fully attributable to more diverse rollout sampling, but arises from the smoother policy distribution itself as a better initialization for continued optimization.
As a supplementary evaluation beyond mathematical reasoning, PR-FKL also yields modest improvements in a CodeContest-trained setting, with full results reported in ~\cref{app:code_domain}.

\subsection{Effect of Policy Reheating on Model Internals}

To understand what high-temperature OPSD changes in the model, we analyze its effects at three levels of increasing proximity to the output: internal representations, output distributions, and reasoning capability.
For tractability of layer-wise analysis, representation and output distribution experiments in this section are conducted on Qwen3-0.6B, whose shallower architecture (28 layers) makes representational changes easier to inspect. 
The reasoning capability analysis uses the entropy-collapsed Qwen3-8B-Base checkpoint, consistent with the main experiments.

\begin{figure}[!t]
    \centering
    \includegraphics[width=\linewidth]{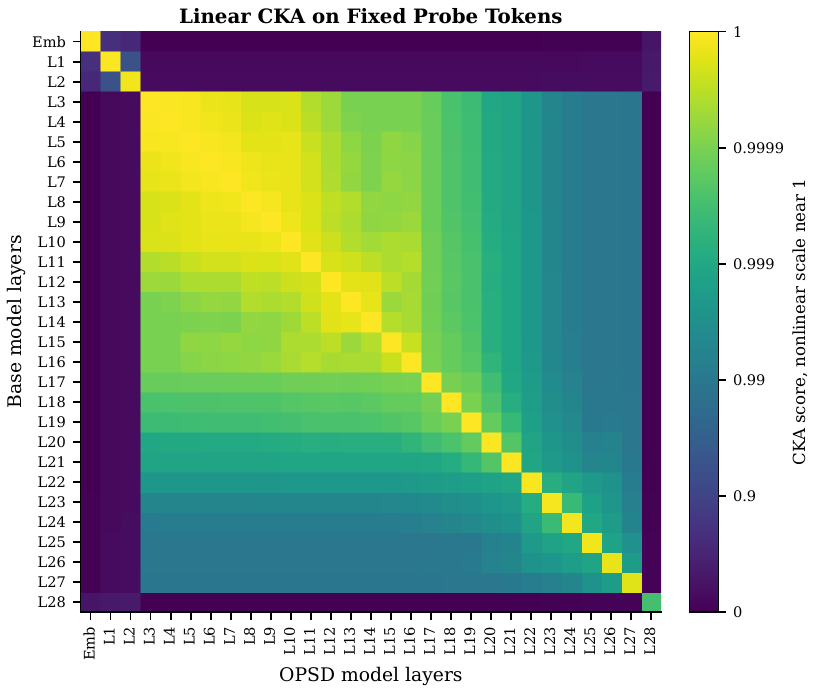}
    \caption{
        Linear CKA between layers of the base and OPSD model on fixed probe tokens.
        The colorbar uses a nonlinear scale near 1 to resolve fine differences.
        Diagonal entries from all layers are near 1, indicating that intermediate-layer representations are essentially unchanged after OPSD.
        The final layer L28 shows a measurable but still small deviation, suggesting that parameter changes are concentrated near the output.
    }
    \label{fig:cka}
\end{figure}

\paragraph{Representation analysis.}
~\cref{fig:cka} shows the linear CKA matrix between every pair of layers across the collapsed and OPSD-reheated model.
All diagonal entries fall above 0.999 on the nonlinear colorbar scale, indicating that internal representations are essentially unchanged throughout the network after reheating.
The deviation is slightly larger in deeper layers (L25--L28), with L28 showing the largest relative drop, suggesting that what little change OPSD induces is concentrated near the output projection rather than in the intermediate layers that encode semantic and reasoning structure.

\begin{figure*}[!t]
    \centering
    \begin{subfigure}[t]{0.32\linewidth}
        \includegraphics[width=\linewidth]{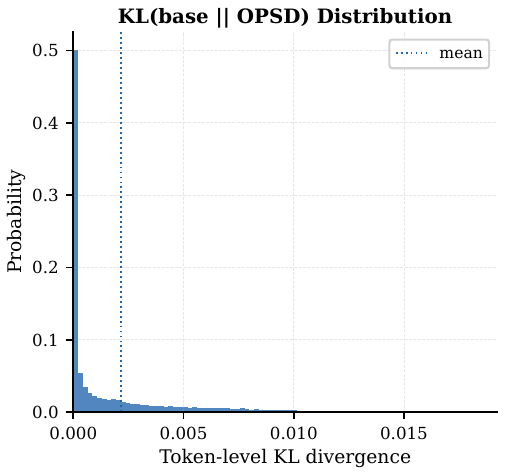}
        \caption{Token-level KL divergence distribution.}
        \label{fig:kl_dist}
    \end{subfigure}
    \hfill
    \begin{subfigure}[t]{0.32\linewidth}
        \includegraphics[width=\linewidth]{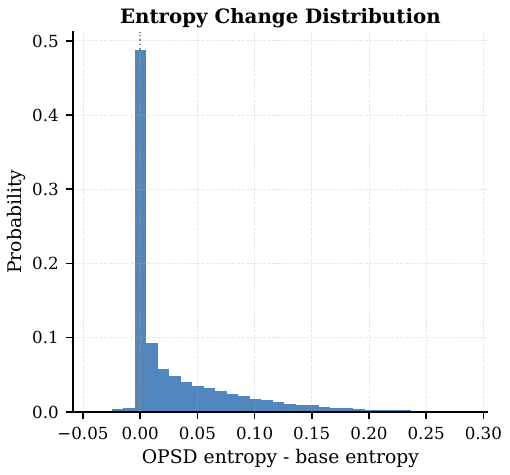}
        \caption{Per-token entropy change distribution.}
        \label{fig:entropy_change}
    \end{subfigure}
    \hfill
    \begin{subfigure}[t]{0.32\linewidth}
        \includegraphics[width=\linewidth]{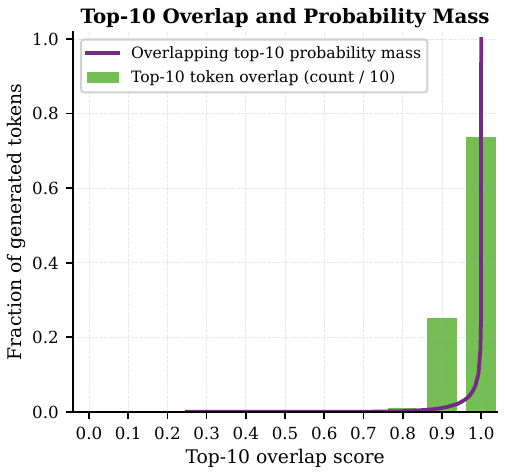}
        \caption{Top-10 token overlap and probability mass retention.}
        \label{fig:top10}
    \end{subfigure}
    \caption{
        Output distribution analysis before and after high-temperature OPSD on MATH-500 probe data.
        Left: token-level KL divergence is concentrated near zero (mean $\approx$ 0.002 nats), indicating a small absolute shift.
        Center: entropy change is systematically positive, confirming a consistent increase in output diversity.
        Right: top-10 token overlap and overlapping probability mass are both concentrated near 1, showing that the candidate support set is almost entirely preserved.
    }
    \label{fig:distribution}
\end{figure*}

\paragraph{Output distribution analysis.}
This localized picture extends to the output distribution level.
~\cref{fig:distribution} shows three complementary views of how the output distribution changes after OPSD.
The token-level KL divergence (\cref{fig:kl_dist}) is heavily concentrated near zero with a mean of approximately 0.002 nats, and the top-10 candidate set is almost entirely preserved (\cref{fig:top10}): both token overlap rate and overlapping probability mass are concentrated near 1, confirming that the model's semantic preferences and candidate support are left intact.
The one meaningful change is entropy: the per-token entropy change distribution (\cref{fig:entropy_change}) is systematically right-skewed, indicating a consistent and directed increase in output diversity across token positions.
That is, high-temperature OPSD performs \textit{support-preserving probability redistribution}; it selectively softens top-candidate margins without adding tokens or shifting semantics; this targeted entropy gain distinguishes it from the near-zero changes at parameter and support levels.

\begin{figure}[!t]
    \centering
    \includegraphics[width=\linewidth]{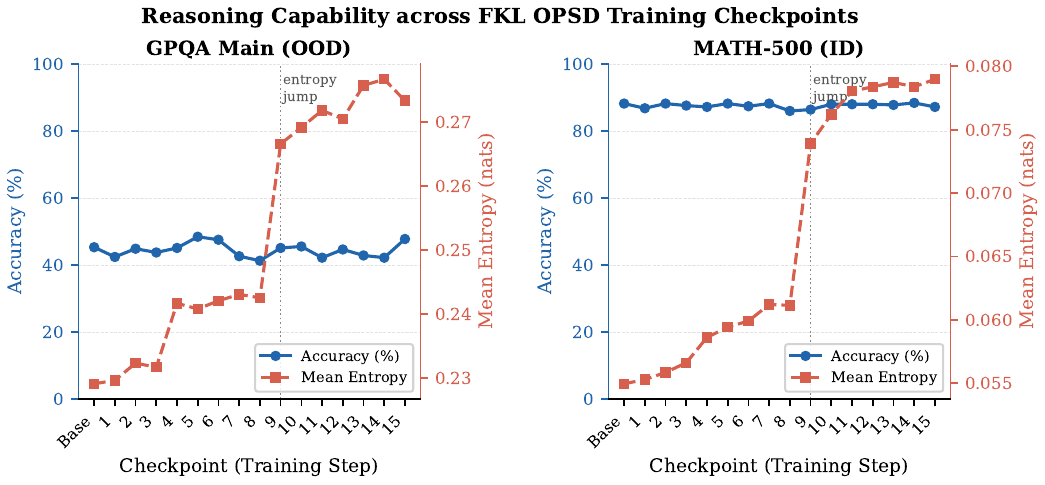}
    \caption{
        Task accuracy and mean token entropy across FKL OPSD training checkpoints
        on MATH-500 (ID) and GPQA Main (OOD).
        The dashed vertical line marks the entropy jump.
        Accuracy on both benchmarks remains stable throughout training,
        confirming that entropy recovery does not degrade reasoning capability.
    }
    \label{fig:reasoning}
\end{figure}

\paragraph{Reasoning capability.}
The conservative nature of these changes is further reflected in downstream reasoning performance.
~\cref{fig:reasoning} plots task accuracy on MATH-500 and GPQA Main~\citep{rein2023gpqagraduatelevelgoogleproofqa} alongside mean token entropy across intermediate OPSD checkpoints.
Accuracy on both benchmarks remains within normal evaluation variance throughout distillation, with no systematic degradation observed even as entropy increases steadily, confirming that OPSD modifies output sharpness without disrupting the reasoning capabilities acquired during RL training.

\subsection{Training Dynamics and Entropy Jump Phenomenon}

\paragraph{Entropy jump phenomenon.}
As shown in ~\cref{fig:reasoning}, entropy growth during OPSD training is not uniform but exhibits a characteristic two-phase dynamic: a slow, gradual increase followed by a sharp discontinuous jump, after which entropy stabilizes at a significantly higher level.
We hypothesize that these two phases reflect qualitatively different mechanisms.
During the slow phase, OPSD smooths the margin between existing top candidates without substantially changing sampling behavior --- the policy is being locally flattened, but rollouts remain on familiar reasoning paths.
The jump, by contrast, reflects a phase transition: certain low-probability reasoning paths cross the sampling threshold and enter the on-policy rollout distribution for the first time, and due to autoregressive causal dependence, these new prefixes drive subsequent token distributions into previously unvisited regions, producing the abrupt entropy increase.
Crucially, the entropy jump is not accompanied by any degradation in reasoning performance: accuracy on both MATH-500 and GPQA Main is maintained after the transition (\cref{fig:reasoning}), indicating that the newly explored paths are productive rather than degenerate.

\begin{figure}[t]
    \centering
    \includegraphics[width=\linewidth]{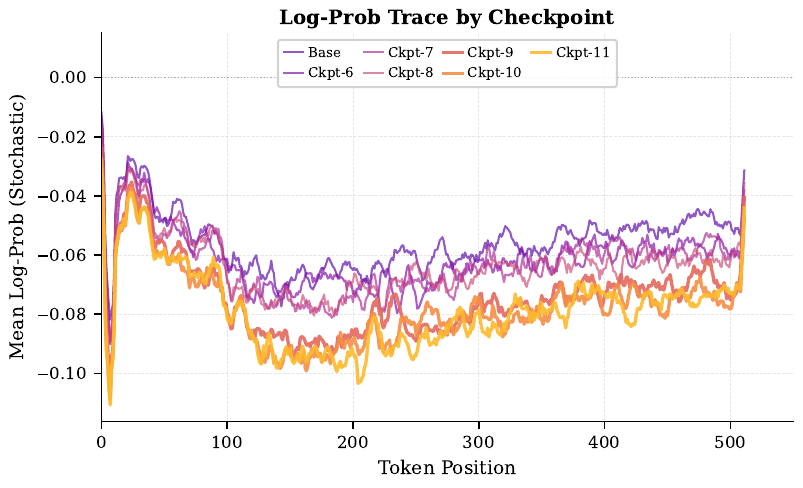}
    \caption{
    Stochastic log-probability trace averaged across token positions for each FKL OPSD checkpoint on MATH-500 probe data.
    Pre-jump checkpoints (Base, Ckpt-6/7/8) show a gradual, undifferentiated decline across all positions with no clear inter-checkpoint separation.
    Post-jump checkpoints (Ckpt-9/10/11) exhibit markedly lower mean log-probabilities beginning at approximately token position 50, with clear stratification persisting through the remainder of the sequence, indicating that the model begins to explore lower-probability reasoning branches more frequently after the entropy jump.
    }
    \label{fig:lp_trace}
\end{figure}

\paragraph{Evidence for the causal-dependence hypothesis.}
Two token-level metrics provide evidence consistent with this picture.
The stochastic log-probability trace (\cref{fig:lp_trace}) shows that pre-jump checkpoints (Base, Ckpt-6/7/8) exhibit only a gradual and largely undifferentiated decline in mean log-probability across token positions.
In contrast, post-jump checkpoints (Ckpt-9/10/11) show a steeper decline beginning around token position 50, with the separation persisting through the remainder of the sequence.
This positional pattern is consistent with the causal-dependence hypothesis: the distributional shift emerges not at the prompt-processing stage, but near the point where reasoning trajectories begin to unfold.

~\cref{tab:diversity} provides a complementary view based on stochastic rollouts compared against greedy-decoded references.
All three statistics exhibit a synchronized change around the entropy jump.
The mean fork position remains around 57 before the jump but drops to approximately 49--50 afterward, closely matching the token-50 onset observed in ~\cref{fig:lp_trace}.
At the same transition, the absolute mean log-probability increases from about 0.061 at Ckpt-8 to 0.074 at Ckpt-9, indicating more frequent sampling of lower-probability continuations.
Edit distance shows a smaller but still consistent shift.
The joint shift in fork position, log-probability, and edit distance suggests that the entropy jump corresponds to a change in rollout behavior: stochastic generations begin to branch earlier and follow more distinct, lower-probability reasoning trajectories, rather than merely exhibiting uniformly distributed surface variation.
Together, these results support viewing the entropy jump as a transition from local probability smoothing to active exploration of lower-probability reasoning branches.

\begin{table}[t]
\centering
\small
\setlength{\tabcolsep}{8pt}
\begin{tabular}{lccc}
\toprule
\textbf{Checkpoint} & \textbf{Edit Dist.} & \textbf{Fork Pos.} & \textbf{$|$Log-Prob$|$} \\
\midrule
Base     & 0.581 & 61.7 & 0.0545 \\
Ckpt-6   & 0.591 & 56.4 & 0.0597 \\
Ckpt-7   & 0.594 & 57.3 & 0.0619 \\
Ckpt-8   & 0.591 & 56.8 & 0.0613 \\
\rowcolor{orange!12}
Ckpt-9   & 0.601 & 50.3 & 0.0738 \\
\rowcolor{orange!12}
Ckpt-10  & 0.601 & 48.4 & 0.0767 \\
\rowcolor{orange!12}
Ckpt-11  & 0.605 & 49.7 & 0.0774 \\
\bottomrule
\end{tabular}
\caption{
Diversity and log-probability statistics across OPSD checkpoints.
\textbf{Edit Dist.}: mean character-level edit distance between each stochastic rollout and a greedy-decoded reference sequence.
\textbf{Fork Pos.}: mean token position at which a stochastic rollout first diverges from the greedy reference (lower = earlier branching).
\textbf{$|$Log-Prob$|$}: absolute mean log-probability under stochastic sampling (higher = more low-probability exploration).
Shaded rows: post-entropy-jump checkpoints.
}
\label{tab:diversity}
\end{table}

\subsection{Analysis of KL Objectives and Temperature Sensitivity}
\label{sec:kl_analysis}

\paragraph{Gradient structure under self-temperature coupling.}
A distinctive property of high-temperature OPSD is that its teacher is not an arbitrary external distribution, but a monotone power transform of the student itself:
$p_T(v) \propto q_\theta(v)^{1/T}$.
This coupling preserves the student's token ranking and largely avoids the head-tail mismatch that can arise in standard KD, where the teacher and student may assign high probability to different regions of the vocabulary.
It also gives the two KL objectives different interpretations.
We provide full derivations in ~\cref{app:kl_derivation} and summarize the key results here.

For FKL, the logit gradient takes the form $q_\theta(v) - p_T(v)$.
The update, therefore, moves the student toward a temperature-smoothed version of its own distribution: it compresses logit margins while preserving semantic preference ordering.
We refer to this as \textit{structured smoothing}: the update re-weights suppressed alternatives according to the model's own prior judgments rather than imposing an externally determined redistribution.

For RKL, the derivation in ~\cref{app:kl_derivation} yields a particularly simple form:
\begin{equation}
\nabla_z \mathcal{L}_{\mathrm{RKL}} = -\!\left(1 - \frac{1}{T}\right)\nabla_z H(q_\theta).
\end{equation}
Thus, under the self-temperature coupling, minimizing RKL is locally equivalent to maximizing the student's entropy, with temperature acting only as a scalar multiplier on the gradient magnitude.
Unlike FKL, RKL does not explicitly match the student to a particular high-temperature target distribution, but instead induces a more generic entropy-increasing update with direction independent of $T$.

\begin{figure}[t]
    \centering
    \includegraphics[width=\linewidth]{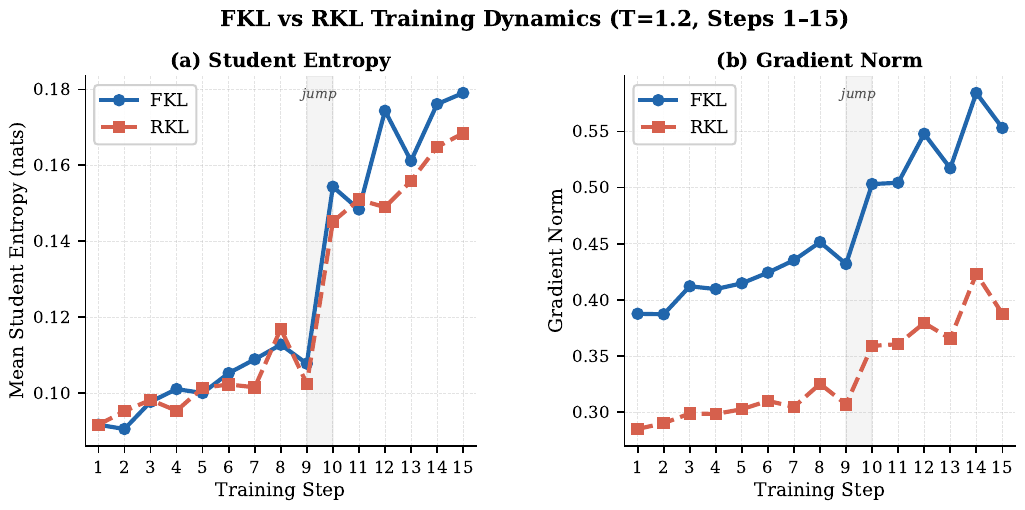}
    \caption{
    Training dynamics of FKL and RKL objectives during high-temperature OPSD ($T=1.2$, steps 1--15).
    (a)~Mean student entropy: both objectives produce comparable entropy recovery, with a jump occurring around step 9--10 under both settings.
  (b)~Gradient norm: FKL consistently exhibits a higher gradient norm than RKL throughout training, consistent with the theoretical prediction that RKL reduces to a more conservative entropy-maximizing update with a smaller gradient magnitude.
    }
    \label{fig:fkl_rkl_compare}
\end{figure}

\paragraph{Empirical dynamics and implications for continued RL.}
Despite this theoretical asymmetry, both objectives recover entropy to a similar degree in practice (\cref{fig:fkl_rkl_compare}).
FKL consistently exhibits a larger gradient norm throughout training, while RKL achieves comparable entropy recovery with smaller gradients, consistent with its interpretation as a direct entropy-maximizing update.
However, comparable entropy recovery does not imply that the resulting checkpoints are equally useful for continued RL.
The distinction above offers a natural explanation for the performance gap between PR-FKL and PR-RKL observed in ~\cref{sec:main_results}.
FKL broadens the distribution by re-weighting alternatives according to the model's existing preference geometry, which helps preserve the reasoning structure acquired during prior RL.
RKL, in contrast, is closer to entropy equalization: it raises entropy across the support without explicitly preserving the relative merit of different reasoning branches.

\paragraph{Temperature sensitivity and applicable scope.}
For RKL, temperature insensitivity follows immediately from the gradient formula, since $T$ enters only as the scalar $(1-1/T)$ and affects only the effective learning rate.
For FKL, insensitivity is not guaranteed a priori, yet empirically FKL also exhibits negligible sensitivity across $T \in \{1.2, 1.5, 2.0\}$ on entropy-collapsed checkpoints (\cref{fig:temperature_entropy_fkl} in ~\cref{app:temperature_analysis}), with entropy growth trajectories converging to within 0.01 nats.
The natural explanation is that collapsed policies have extremely sharp logit distributions, causing teachers at different temperatures to induce nearly parallel gradient directions.
Notably, this insensitivity is itself a consequence of the collapsed regime: on models with already-elevated entropy, the self-teacher rapidly approaches a near-uniform distribution, KL gradients grow large, and entropy diverges within a few steps (see ~\cref{app:temperature_analysis}).
High-temperature OPSD should therefore be understood as a post-collapse reheating tool, gated by the current entropy level of the policy.

\section{Conclusion}

We presented TS-OPSD, a lightweight policy reheating method that internalizes the exploratory effect of temperature into model parameters, rather than treating it as an external decoding or rollout control. 
Applied as a post-collapse intervention, TS-OPSD requires no external teacher, privileged data, or extra inference cost.

Empirically, reheating an entropy-collapsed checkpoint provides a stronger initialization for continued RL than standard continued training, entropy-preserving RL alternatives, and rollout-level temperature reheating.
Our analyses suggest that this improvement comes from a conservative form of policy smoothing: entropy and rollout diversity increase while candidate support, intermediate representations, and downstream reasoning ability are largely preserved.

More broadly, our results show that entropy restoration can be decoupled from the RL objective and inserted as a modular inter-stage operation when the policy becomes too deterministic. 
This perspective points to RL training pipelines in which exploitation and exploration recovery are handled by separate but complementary stages, offering a simple way to extend reasoning-oriented RL beyond the entropy-collapse bottleneck.

\section*{Limitations}

Due to computational constraints, our main experiments are limited to Qwen3-4B-Base and Qwen3-8B-Base, and we have not validated TS-OPSD on larger-scale models or other model families.
Although the CodeContest experiment in \cref{app:code_domain} provides initial evidence beyond mathematical RLVR, we do not evaluate tool use or other reward settings.
The effectiveness of policy reheating may therefore vary with model family, scale, task domain, and reward design, and broader empirical coverage is needed to establish generality.

Furthermore, the end-to-end evaluation pipeline that combines OPSD and continued RL is computationally expensive, which precludes a systematic grid search over key hyperparameters.
Specifically, the entropy collapse threshold and the entropy recovery target are set empirically based on observed training dynamics rather than through principled optimization.
While the dynamics-based recipe in \cref{sec:exp_setup} is more transferable than the absolute values, the rules have not been systematically tuned across settings; identifying more robust or adaptive criteria remains an open direction for future work.

Finally, our experiments consider only a single round of policy reheating followed by continued RL.
A natural next step is to study cascaded schedules that alternate OPSD and RL across multiple stages, using policy reheating to repeatedly restore exploration as entropy collapse reappears during longer training runs.
However, such multi-round pipelines introduce additional sources of variation, including when to trigger each reheating intervention and how many rounds are beneficial.
Since our current trigger and stopping criteria are still empirically chosen rather than principled, a multi-round study would substantially increase both computational cost and hyperparameter sensitivity.
To avoid conflating the effect of policy reheating with these additional scheduling choices, we focus on a single-intervention setting and leave principled cascaded OPSD--RL schedules for future work.

\section*{Acknowledgments}

This work was partly supported by the National Natural Science Foundation of China (Grant No.~62576191) and the Tsinghua SIGS KA Cooperation Fund.

\bibliography{custom}

\clearpage
\appendix

\section{Hyperparameters}
\label{app:hyperparams}

\subsection{RL Training (DAPO)}
\label{app:hyperparams:rl}

We use DAPO \citep{yu2025dapoopensourcellmreinforcement} as the base RL procedure for all experiments, implemented on top of verl \citep{sheng2024hybridflow}. 
~\cref{tab:hyperparams_rl} lists the shared hyperparameters for Qwen3-4B-Base and Qwen3-8B-Base; the two models use identical settings unless otherwise noted.

\begin{table}[h!]
\centering
\begin{tabular}{ll}
\toprule
\textbf{Hyperparameter} & \textbf{Value} \\
\midrule
Framework              & verl \\
Training data          & DAPO-Math-17K \\
Optimizer              & AdamW \\
Learning rate          & 2e-6 \\
LR warmup steps        & 10 \\
Weight decay           & 0.1 \\
Train batch size       & 64 \\
Responses per prompt   & 8 \\
PPO mini-batch size    & 64 \\
Clip ratio low         & 0.2 \\
Clip ratio high        & 0.28 \\
Clip ratio $c$         & 10.0 \\
Gradient clip          & 1.0 \\
Loss aggregation       & token-mean \\
KL loss                & disabled \\
Entropy coefficient    & 0.0 \\
Max response length    & 8192 \\
Rollout temperature    & 1.0 \\
Rollout top-$p$        & 1.0 \\
Overlong buffer        & enabled, length 1024 \\
Total epochs           & 2 \\
\bottomrule
\end{tabular}
\caption{Hyperparameters for DAPO RL training.}
\label{tab:hyperparams_rl}
\end{table}

For DAPO w/ RR (Rollout Reheating), all hyperparameters are identical 
except the rollout sampling temperature, which is raised to $T=1.2$ when 
resuming from the collapsed checkpoint.

\subsection{OPSD (Policy Reheating)}
\label{app:hyperparams:opsd}

OPSD is implemented via the GKD framework in swift \citep{zhao2025swiftascalablelightweightinfrastructure}. 
The teacher distribution is constructed from the student's own logits via temperature scaling with stop-gradient, and the student is updated using full-vocabulary KL with on-policy rollouts at each step. 
~\cref{tab:hyperparams_opsd} lists the training hyperparameters; PR-FKL and PR-RKL share all settings and differ only in KL direction.
For Qwen3-4B-Base, the reheated checkpoint used to resume RL is selected at step 6, at which point mean token entropy has recovered above 0.12 nats; for Qwen3-8B-Base, whose entropy growth rate under OPSD is slower, the checkpoint at step 15 is used instead.

\begin{table}[h]
\centering
\begin{tabular}{ll}
\toprule
\textbf{Hyperparameter} & \textbf{Value} \\
\midrule
Framework              & GKD (swift) \\
Teacher temperature $T$ & 1.2 \\
Logit coverage         & Full vocabulary \\
Learning rate          & 1e-6 \\
LR schedule            & Constant \\
Effective batch size   & 1024 \\
Rollout sampling       & On-policy, $T=1.0$ \\
Precision              & bfloat16 \\
Max completion length  & 4096 \\
\bottomrule
\end{tabular}
\caption{Hyperparameters for high-temperature OPSD.}
\label{tab:hyperparams_opsd}
\end{table}

\subsection{Evaluation}
\label{app:hyperparams:eval}

All models are evaluated using VLLM for generation, followed by rule-based answer extraction and exact-match scoring. 
For OlymMATH, we evaluate on the English subset only (easy and hard splits combined), excluding the Chinese and Lean formalization variants.
~\cref{tab:hyperparams_eval} lists the decoding hyperparameters applied uniformly across all benchmarks and models.

\begin{table}[h]
\centering
\begin{tabular}{ll}
\toprule
\textbf{Hyperparameter} & \textbf{Value} \\
\midrule
Samples per prompt ($n$) & 8 \\
Sampling temperature   & 1.0 \\
Top-$p$                & 0.7 \\
Top-$k$                & disabled \\
Max new tokens         & 8192 \\
Precision              & bfloat16 \\
\bottomrule
\end{tabular}
\caption{Hyperparameters for evaluation.}
\label{tab:hyperparams_eval}
\end{table}

\section{Derivation of KL Gradients under Self-Temperature Teacher}
\label{app:kl_derivation}

Let $z \in \mathbb{R}^{|V|}$ denote the student logits and
$q(v)=\mathrm{softmax}(z)_v$ the student distribution. 
Given a temperature $T>1$, define
\begin{equation}
\alpha=\frac{1}{T}, 
\qquad 
Z_\alpha=\sum_u q(u)^\alpha .
\end{equation}
The self-temperature teacher is
\begin{equation}
p_T(v)
=
\frac{\exp(z_v/T)}{\sum_u \exp(z_u/T)}
=
\frac{q(v)^\alpha}{Z_\alpha},
\label{eq:self_temp_teacher}
\end{equation}
where $p_T$ is treated as stop-gradient during the distillation update.

\paragraph{Forward KL.}
The forward KL objective is
\begin{equation}
\mathcal{L}_{\mathrm{FKL}}
=
D_{\mathrm{KL}}(p_T\|q)
=
\sum_v p_T(v)\log\frac{p_T(v)}{q(v)} .
\end{equation}
Since $p_T$ is stop-gradient, only the $\log q(v)$ term contributes to the gradient. 
Using the standard softmax log-gradient identity
\begin{equation}
\frac{\partial \log q(u)}{\partial z_v}
=
\mathbf{1}\{u=v\}-q(v),
\label{eq:softmax_log_grad}
\end{equation}
we have
\begin{equation}
\begin{aligned}
\frac{\partial \mathcal{L}_{\mathrm{FKL}}}{\partial z_v}
&=
-\sum_u p_T(u)
\frac{\partial \log q(u)}{\partial z_v}  \\
&=
-\sum_u p_T(u)
\bigl(\mathbf{1}\{u=v\}-q(v)\bigr) \\
&=
-p_T(v)+q(v)\sum_u p_T(u) \\
&=
q(v)-p_T(v).
\end{aligned}
\label{eq:fkl_grad}
\end{equation}
Thus, gradient descent updates the logits in the direction
\begin{equation}
\Delta z_v^{\mathrm{FKL}}
\propto
p_T(v)-q(v).
\end{equation}
Moreover,
\begin{equation}
\frac{p_T(v)}{q(v)}
=
\frac{q(v)^{\alpha-1}}{Z_\alpha},
\qquad \alpha<1.
\label{eq:pt_q_ratio}
\end{equation}
Therefore, $p_T(v)/q(v)$ is monotonically decreasing in $q(v)$.
Temperature scaling relatively down-weights high-probability tokens and relatively up-weights lower-probability alternatives, while preserving the logit-induced token ranking.

\paragraph{Reverse KL.}
The reverse KL objective is
\begin{equation}
\mathcal{L}_{\mathrm{RKL}}
=
D_{\mathrm{KL}}(q\|p_T)
=
\sum_u q(u)\log\frac{q(u)}{p_T(u)} .
\end{equation}
We first derive the gradient for a fixed stop-gradient teacher $p_T$.
Using the softmax derivative
\begin{equation}
\frac{\partial q(u)}{\partial z_v}
=
q(u)\bigl(\mathbf{1}\{u=v\}-q(v)\bigr),
\label{eq:softmax_grad}
\end{equation}
we obtain
\begin{equation}
\begin{aligned}
\frac{\partial \mathcal{L}_{\mathrm{RKL}}}{\partial z_v}
&=
\sum_u
\frac{\partial q(u)}{\partial z_v}
\left(
\log\frac{q(u)}{p_T(u)}+1
\right) \\
&=
\sum_u
q(u)\bigl(\mathbf{1}\{u=v\}-q(v)\bigr) \\
&\qquad\cdot\left(
\log\frac{q(u)}{p_T(u)}+1
\right) \\
&=
q(v)\left(
\log\frac{q(v)}{p_T(v)}+1
\right) \\
&\quad
-q(v)\sum_u q(u)
\left(
\log\frac{q(u)}{p_T(u)}+1
\right) \\
&=
q(v)\left(
\log\frac{q(v)}{p_T(v)}
-
\mathcal{L}_{\mathrm{RKL}}
\right).
\end{aligned}
\label{eq:rkl_general_grad}
\end{equation}

We now specialize Eq.~\eqref{eq:rkl_general_grad} to the self-temperature teacher in Eq.~\eqref{eq:self_temp_teacher}. At the point where $p_T$ is constructed,
\begin{equation}
\log\frac{q(v)}{p_T(v)}
=
(1-\alpha)\log q(v)+\log Z_\alpha .
\label{eq:rkl_log_ratio}
\end{equation}
Moreover,
\begin{equation}
\begin{aligned}
\mathcal{L}_{\mathrm{RKL}}
&=
\sum_v q(v)
\bigl((1-\alpha)\log q(v)+\log Z_\alpha\bigr) \\
&=
(1-\alpha)\sum_v q(v)\log q(v)+\log Z_\alpha \\
&=
-(1-\alpha)H(q)+\log Z_\alpha .
\end{aligned}
\label{eq:rkl_value}
\end{equation}
Substituting Eq.~\eqref{eq:rkl_log_ratio} and Eq.~\eqref{eq:rkl_value} into Eq.~\eqref{eq:rkl_general_grad}, the $\log Z_\alpha$ terms cancel:
\begin{equation}
\begin{aligned}
\frac{\partial \mathcal{L}_{\mathrm{RKL}}}{\partial z_v}
&=
q(v)\Bigl[
(1-\alpha)\log q(v)+\log Z_\alpha \\
&\qquad\quad
+(1-\alpha)H(q)-\log Z_\alpha
\Bigr] \\
&=
(1-\alpha)q(v)\bigl(\log q(v)+H(q)\bigr).
\end{aligned}
\label{eq:rkl_grad}
\end{equation}

Finally, the entropy gradient of $q$ with respect to logits is
\begin{equation}
\frac{\partial H(q)}{\partial z_v}
=
-q(v)\bigl(\log q(v)+H(q)\bigr).
\label{eq:entropy_grad}
\end{equation}
Combining Eq.~\eqref{eq:rkl_grad} and Eq.~\eqref{eq:entropy_grad}, we obtain
\begin{equation}
\begin{aligned}
\nabla_z \mathcal{L}_{\mathrm{RKL}}
&=
-(1-\alpha)\nabla_z H(q) \\
&=
-\left(1-\frac{1}{T}\right)\nabla_z H(q).
\end{aligned}
\label{eq:rkl_entropy_grad}
\end{equation}
Thus, under a stop-gradient self-temperature teacher, minimizing RKL is locally equivalent to maximizing the entropy of the student distribution, with temperature only scaling the gradient magnitude by $1-1/T$.

\section{Analysis of Temperature Sensitivity and Applicable Scope}
\label{app:temperature_analysis}

\begin{figure}[h]
    \centering
    \includegraphics[width=\linewidth]{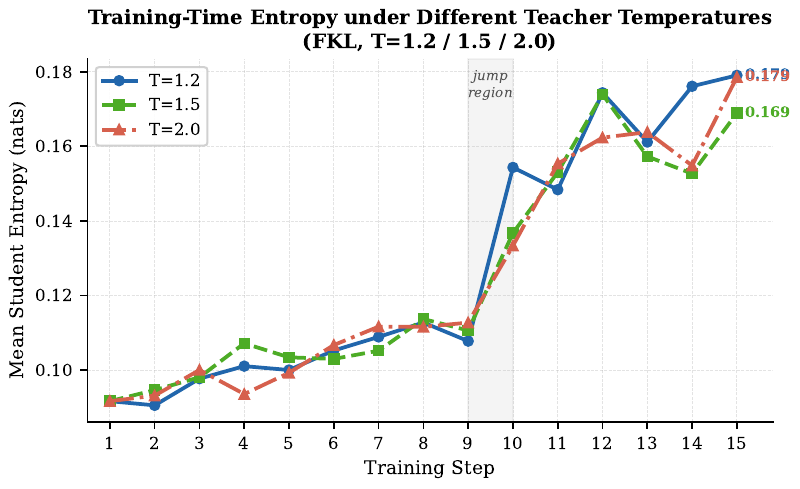}
    \caption{
    Training-time entropy dynamics under different teacher temperatures($T \in \{1.2, 1.5, 2.0\}$) for FKL OPSD on an entropy-collapsed checkpoint.
    All three temperatures produce nearly identical entropy growth trajectories, with the jump region occurring at similar steps and final entropy levels converging to within 0.01~nats.
    This empirical insensitivity is consistent with the observation that collapsed policies have extremely sharp logit distributions, causing high-temperature teachers at different $T$ values to induce nearly parallel gradient directions.
    }
    \label{fig:temperature_entropy_fkl}
\end{figure}

\begin{figure}[h]
    \centering
    \includegraphics[width=\linewidth]{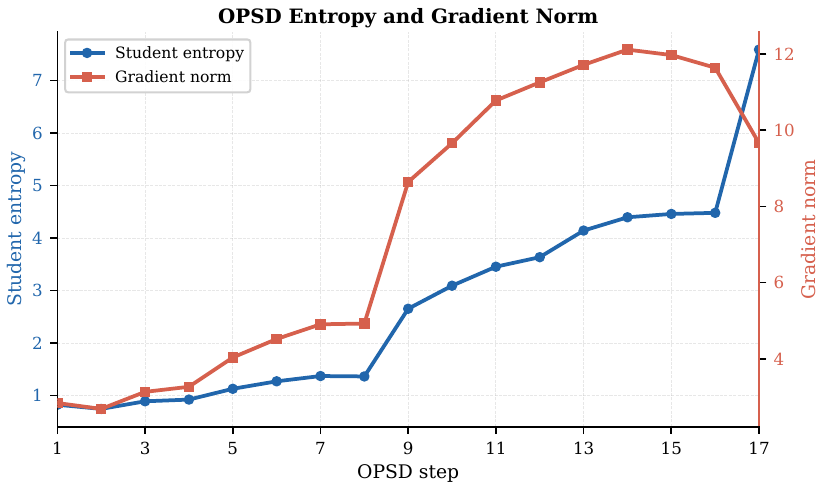}
    \caption{
    Student entropy and gradient norm during FKL OPSD applied to Qwen3-8B-Base \emph{before} any RL training.
    In contrast to the controlled entropy recovery observed on entropy-collapsed checkpoints (\cref{fig:temperature_entropy_fkl}), entropy here rises continuously from approximately 0.8~nats at step~1 to over 7~nats by step~17, far exceeding the target range of the reheating procedure.
    Gradient norm increases sharply from step~9 onward and remains elevated, indicating that training has entered an unstable regime.
    This behavior arises because the base model's already-elevated entropy causes the self-teacher to rapidly approach a near-uniform distribution, producing large forward KL gradients that drive uncontrolled entropy growth.
    }
    \label{fig:base_entropy_diverge}
\end{figure}

~\cref{fig:temperature_entropy_fkl} shows that on entropy-collapsed checkpoints, high-temperature FKL OPSD is robust to teacher temperature across $T \in \{1.2, 1.5, 2.0\}$: all three settings produce nearly identical entropy trajectories, and the method converges reliably to the target entropy range.
~\cref{fig:base_entropy_diverge} illustrates what happens when the same procedure is applied outside this regime, to a pretrained base model before any RL training.
Entropy rises continuously and without bound, and gradient norm grows sharply from step~9 onward, indicating an unstable training dynamic. 
The contrast between the two settings confirms that the stability of high-temperature OPSD is contingent on the policy being in a low-entropy collapsed state: it is precisely the sharpness of the collapsed distribution that keeps the self-teacher's gradient directions nearly parallel and the entropy recovery controlled.
High-temperature OPSD should therefore be understood as a post-collapse reheating tool and should not be applied to models with already-elevated entropy.

\section{Long-Horizon OPSD on Entropy-Collapsed Models}
\label{app:long_horizon_opsd}

\begin{figure}[h]
    \centering
    \includegraphics[width=\linewidth]{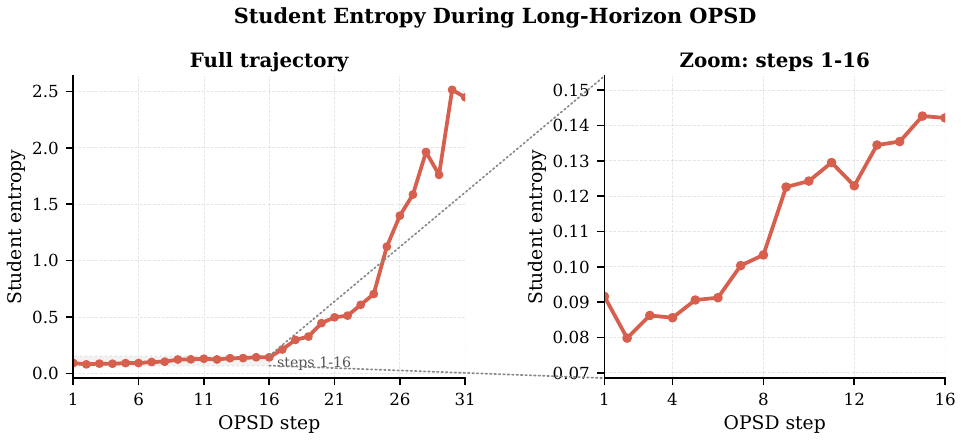}
    \caption{Student entropy during long-horizon FKL OPSD on an entropy-collapsed Qwen3-8B-Base checkpoint. 
    Left: full trajectory over 31 steps, showing a clear regime change around step 16 where entropy transitions from slow, controlled growth to rapid, accelerating increase. 
    Right: zoomed view of steps 1--16, showing the characteristic two-phase dynamic with a visible entropy jump, albeit with higher step-to-step variance due to the reduced effective batch size (256) used in this experiment.}
    \label{fig:long_horizon_opsd}
\end{figure}

~\cref{fig:long_horizon_opsd} presents a long-horizon OPSD experiment on the entropy-collapsed Qwen3-8B-Base checkpoint, extending training well beyond the stopping criterion used in the main experiments. 
The full trajectory (left) reveals a sharp regime change around step 16: before this point, entropy grows slowly and remains within a controlled range (below 0.15 nats), consistent with the reheating dynamics described in ~\cref{sec:kl_analysis}; after step 16, entropy accelerates rapidly and continuously, reaching approximately 2.5 nats by step 31. 
This transition closely mirrors the behavior observed when applying OPSD to a non-collapsed base model (\cref{app:temperature_analysis}): once entropy rises sufficiently above the collapsed regime, the self-teacher approaches a near-uniform distribution, forward KL gradients grow large, and the training dynamic becomes uncontrolled. 
The zoomed view of steps 1--16 (right) shows that the characteristic two-phase dynamic and entropy jump observed in the main analysis remain visible, though with higher step-to-step variance due to the reduced effective batch size (256) used in this experiment, which increases sensitivity to the difficulty distribution of sampled prompts. 
Together, the two panels reinforce the conclusion that high-temperature OPSD is naturally self-limiting when stopped at the appropriate entropy level, but transitions into an unstable regime if training is continued past that point---providing further empirical grounding for the entropy-based stopping criterion described in ~\cref{sec:exp_setup}.

\section{Low-Temperature OPSD as an Approximate Inverse of Reheating}
\label{app:low_temp_opsd}

\begin{figure}[h]
    \centering
    \includegraphics[width=\linewidth]{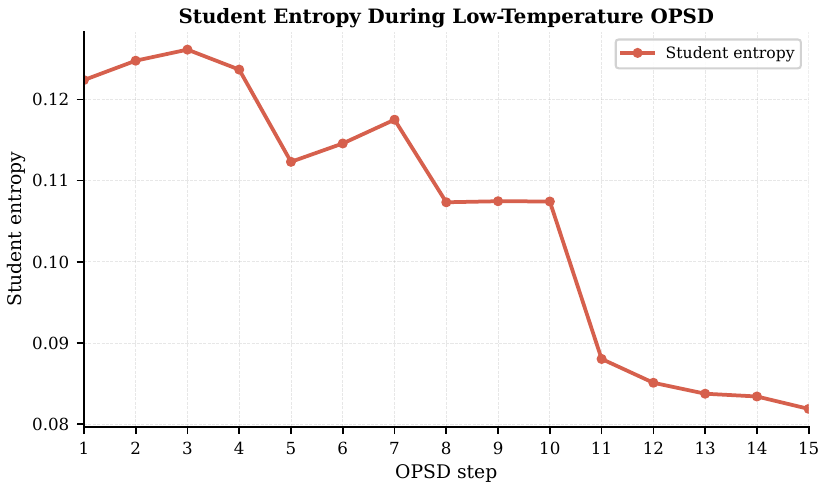}
    \caption{Student entropy during low-temperature ($T=0.8$) RKL OPSD applied to the reheated Qwen3-8B-Base checkpoint. Entropy decreases in a multi-phase pattern symmetric to the jump-and-stabilize dynamics observed during high-temperature OPSD, with a sharp discontinuous drop around step 10--11 after an initial slow decline.}
    \label{fig:lowT_opsd_entropy}
\end{figure}

\begin{figure}[h]
    \centering
    \includegraphics[width=\linewidth]{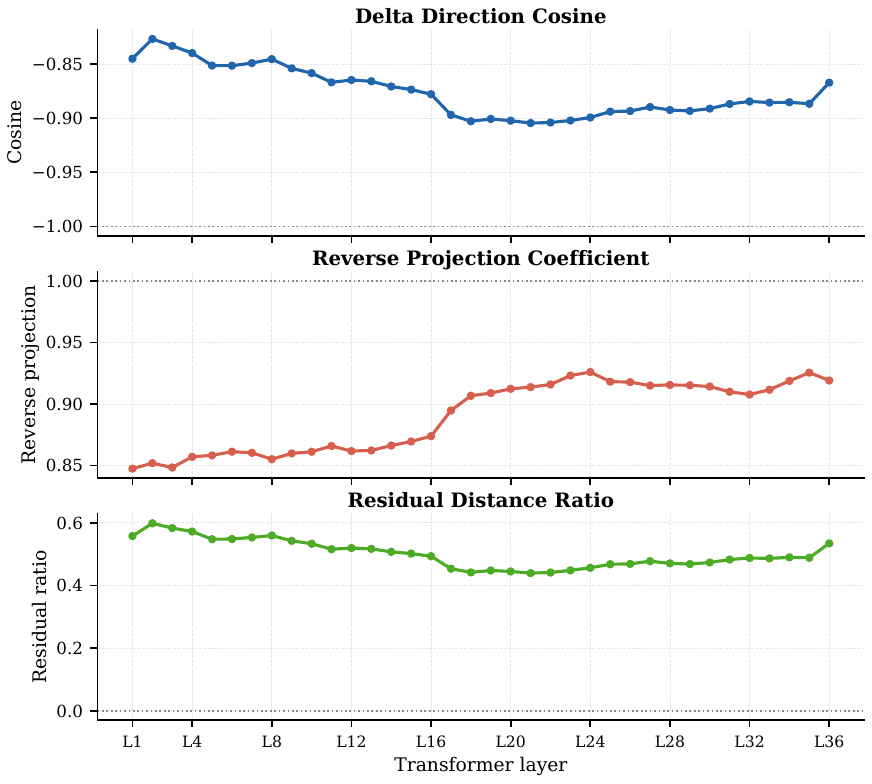}
    \caption{Layer-wise reversibility metrics comparing the parameter update induced by high-temperature OPSD ($\Delta_\text{high} = \theta_\text{reheat} - \theta_\text{collapse}$) and the subsequent low-temperature OPSD update ($\Delta_\text{low} = \theta_\text{re-collapse} - \theta_\text{reheat}$). Top: cosine similarity between $\Delta_\text{low}$ and $-\Delta_\text{high}$, measuring directional anti-alignment. Middle: reverse projection coefficient, measuring the fraction of $\|\Delta_\text{high}\|$ recovered by the projection of $\Delta_\text{low}$ onto $-\Delta_\text{high}$. Bottom: residual distance ratio $\|\theta_\text{re-collapse} - \theta_\text{collapse}\| / \|\theta_\text{reheat} - \theta_\text{collapse}\|$, measuring how much of the reheating-induced displacement remains after low-temperature OPSD.}
    \label{fig:layerwise_reversibility}
\end{figure}

\begin{figure*}[t]
    \centering
    \begin{minipage}[t]{0.48\linewidth}
        \centering
        \includegraphics[width=\linewidth]{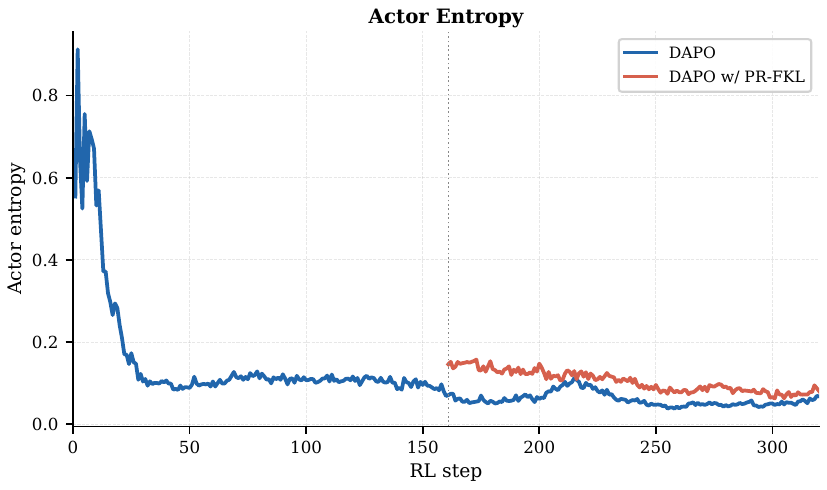}
        \caption*{(a) Actor entropy.}
    \end{minipage}
    \hfill
    \begin{minipage}[t]{0.48\linewidth}
        \centering
        \includegraphics[width=\linewidth]{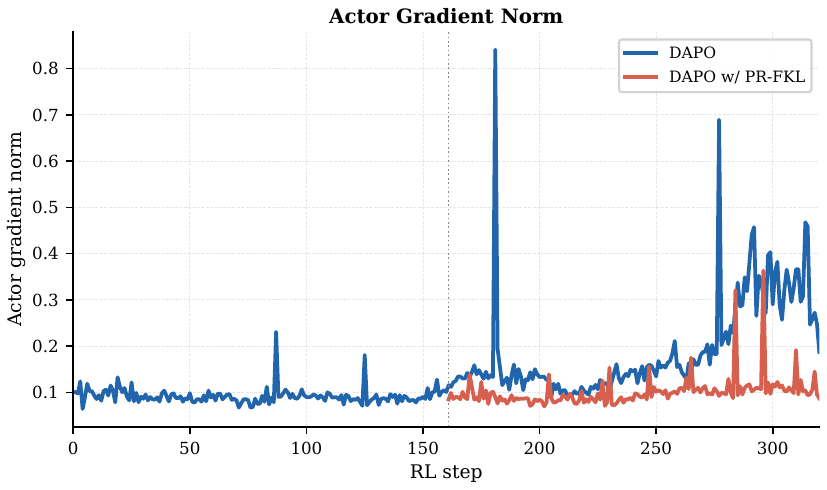}
        \caption*{(b) Actor gradient norm.}
    \end{minipage}

    \vspace{0.5em}

    \begin{minipage}[t]{0.48\linewidth}
        \centering
        \includegraphics[width=\linewidth]{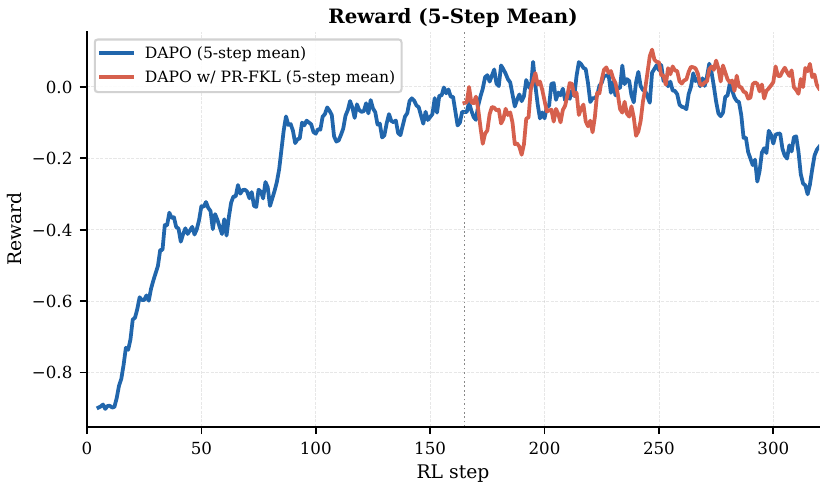}
        \caption*{(c) Reward (5-step mean).}
    \end{minipage}
    \hfill
    \begin{minipage}[t]{0.48\linewidth}
    \end{minipage}

    \caption{Training dynamics on a data-scarce RL setting using a 2048-prompt subset of DAPO-Math-17K (Qwen3-4B-Base).
    Under standard DAPO, entropy collapses to below 0.04 nats early in training, and the policy eventually destabilizes: gradient norm exhibits large spikes and reward deteriorates after approximately step 200.
    After policy reheating (PR-FKL), entropy recovers and remains at a stable, slowly-declining level throughout continued training, with no abnormal gradient norm spikes and no reward degradation.}
    \label{fig:train2048}
\end{figure*}

High-temperature OPSD increases policy entropy by moving model parameters toward a smoother distribution; a natural question is whether low-temperature OPSD---applying RKL distillation toward a sharpened, low-temperature self-teacher---constitutes an approximate inverse of this process at the parameter level.
~\cref{fig:lowT_opsd_entropy} shows the entropy trajectory during low-temperature ($T=0.8$) RKL OPSD applied to the reheated checkpoint. 
Entropy decreases in a multi-phase pattern that mirrors the high-temperature dynamics: unstable fluctuations, followed by a sharp discontinuous drop, and a continued decrease thereafter. 
This structural symmetry suggests that the entropy jump phenomenon is not specific to the reheating direction but reflects a general property of the OPSD training dynamic under self-temperature coupling.

To analyze reversibility at the parameter level, we compute three metrics comparing the high-temperature update vector $\Delta_\text{high} = \theta_\text{reheat} - \theta_\text{collapse}$ and the low-temperature update vector $\Delta_\text{low} = \theta_\text{re-collapse} - \theta_\text{reheat}$, shown in ~\cref{fig:layerwise_reversibility}. 
The global cosine similarity between $\Delta_\text{low}$ and $-\Delta_\text{high}$ is $-0.878$, indicating strong directional anti-alignment that cannot arise by chance in an 8B-parameter space. 
The reverse projection coefficient of $0.891$ shows that the low-temperature update recovers approximately 89\% of the high-temperature displacement in the anti-aligned direction, and the norm ratio of $1.015$ confirms that the two updates are of comparable magnitude. 
However, the residual distance ratio of $0.497$ indicates that the re-collapsed model remains approximately halfway between the original collapsed checkpoint and the reheated checkpoint, rather than returning to the starting point. 
Together, these results are consistent with \textit{partial reversibility}: low-temperature OPSD produces a strongly anti-aligned parameter update that substantially but incompletely reverses the reheating-induced displacement.

The layer-wise breakdown reveals additional structure. 
Directional anti-alignment is consistently stronger in middle and later layers (cosine $\approx -0.90$) than in early layers (cosine $\approx -0.85$), and the reverse projection coefficient rises from approximately $0.85$ in early layers to $0.91$--$0.92$ in deeper layers, suggesting that the reheating update is more faithfully reversed in the layers closer to the output. 
Correspondingly, the residual distance ratio drops to approximately $0.45$ in middle and later layers, indicating higher reversibility there, while early layers retain a larger residual, possibly because shallow representations are less directly constrained by the output-level KL objective.

We note two confounds that prevent a fully controlled interpretation. 
First, the high-temperature and low-temperature OPSD runs were conducted with a decaying rather than constant learning rate, so the effective step sizes at the two stages differ slightly. 
Second, because the low-temperature run starts from the already-modified reheated checkpoint, the gradient field is no longer linear and the low-temperature self-teacher is not the strict mathematical inverse of the high-temperature one. 
These factors likely account for the gap between the observed projection coefficient ($0.891$) and the theoretical maximum of $1.0$, as well as the non-zero residual. 
Despite these limitations, the consistency of the anti-alignment signal across all layers and all three metrics supports the qualitative conclusion that high- and low-temperature OPSD induce structurally opposite parameter movements, and that reheating-induced changes are substantially---though not completely---reversible.

\begin{table*}[!t]
\centering
\small
\setlength{\tabcolsep}{8pt}
\begin{tabular}{lcccc}
\toprule
Model & HE avg@8 & HE pass@8 & HE+ avg@8 & HE+ pass@8 \\
\midrule
Base & 0.723 & 0.945 & 0.654 & 0.896 \\
DAPO & 0.835 & 0.933 & 0.774 & 0.902 \\
DAPO w/ PR-FKL & \textbf{0.843} & \textbf{0.945} & \textbf{0.780} & \textbf{0.915} \\
\bottomrule
\end{tabular}
\caption{Code-domain results after training on CodeContest. PR-FKL shows positive trends over DAPO on all four HumanEval metrics.}
\label{tab:code_domain}
\end{table*}

\section{Policy Reheating under Data-Scarce RL Training}
\label{app:data_scarce}

Post-training compute scaling for LLMs depends not only on model parameters but also on the quality and quantity of training prompts: data scarcity is a genuine bottleneck in RL-based post-training, particularly for larger models that require more diverse and challenging prompts to sustain a meaningful gradient signal. 
To probe the robustness of policy reheating in this regime, we construct a data-scarce setting by restricting RL training to a 2048-prompt subset of DAPO-Math-17K and training Qwen3-4B-Base with the standard DAPO recipe. 
As shown in ~\cref{fig:train2048}, entropy collapse under standard DAPO is substantially more severe in this setting, with the policy entropy dropping below 0.04 nats, reflecting the reduced diversity of on-policy rollouts when the prompt distribution is narrow. 
More critically, training eventually becomes unstable: gradient norm exhibits abnormal spikes around step 200, and reward begins to decline, suggesting that the collapsed policy has entered a regime where further optimization is not only ineffective but actively harmful. 
After applying policy reheating at the collapsed checkpoint and resuming RL, the model maintains a stable, slowly-declining entropy throughout continued training, gradient norm remains well-behaved, and reward does not deteriorate.
These results suggest that policy reheating can serve not only as a performance-enhancing intervention but also as a stabilizing mechanism in data-scarce RL settings, enabling longer, more stable training when prompt diversity is limited.

\section{Code-Domain Experiment}
\label{app:code_domain}

To test whether policy reheating transfers beyond mathematical reasoning, we conduct an additional experiment using CodeContest as the RL training set.
After 120 DAPO steps, mean token entropy decreases to approximately 0.051 nats.
We apply PR-FKL to restore it to 0.127 nats and then resume DAPO under the same configuration.
We evaluate on HumanEval (HE) and HumanEval+ (HE+), reporting avg@8 and pass@8.

As shown in \cref{tab:code_domain}, PR-FKL improves over DAPO on all four metrics, although the gains are smaller than on the mathematical benchmarks.
This is consistent with the motivation that restored exploration is most useful when a non-trivial fraction of tokens are high-entropy branching points that determine a reasoning trajectory; such branching may be less frequent in code generation than in long-form mathematical reasoning.
The result provides initial evidence of cross-domain applicability but should not be interpreted as establishing broad generality.

\section{Computational Cost of Policy Reheating}
\label{app:compute_cost}

Table~\ref{tab:compute_cost} summarizes the computational cost of policy reheating relative to the base RL training, measured on 8$\times$H800 GPUs.
In absolute terms, OPSD requires approximately 4.3 hours for Qwen3-4B-Base and 5.7 hours for Qwen3-8B-Base, compared to over 27.8 and 33.3 hours respectively for the full RL run, placing the overhead of a single reheating intervention at roughly 15--17\% of total RL training time.
When normalizing for batch size---OPSD uses an effective batch size of 1024 versus 64 for RL, a factor of 16$\times$---the per-sample step time of OPSD (approximately 65 seconds for 4B and 86 seconds for 8B) is actually lower than the corresponding RL step time (200 and 240 seconds), indicating that OPSD is not inherently more computationally intensive per sample than RL.
The higher wall-clock time per OPSD step is therefore primarily a consequence of the larger batch size, which is necessary for stable full-vocabulary KL training rather than an intrinsic inefficiency of the method.
Taken together, these figures suggest that policy reheating is a computationally modest intervention: a one-time overhead of roughly 15\% of RL training time that, as shown in the main experiments, yields consistent improvements in continued RL performance.

\clearpage
\onecolumn
\begin{table}[H]
\centering
\small
\setlength{\tabcolsep}{6pt}
\begin{tabular}{lcccc}
\toprule
\multirow{2}{*}{Training stage} & \multicolumn{2}{c}{\textbf{Qwen3-4B-Base}} & \multicolumn{2}{c}{\textbf{Qwen3-8B-Base}} \\
\cmidrule(lr){2-3} \cmidrule(lr){4-5}
& OPSD & RL & OPSD & RL \\
\midrule
Batch size & 1024 & 64 & 1024 & 64 \\
Steps & $\leq$15 & $\geq$500 & $\leq$15 & $\geq$500 \\
Time per step (s) & 1035 & 200 & 1370 & 240 \\
Total time (h) & $\approx$4.3 & $\approx$27.8 & $\approx$5.7 & $\approx$33.3 \\
OPSD / RL total time & \multicolumn{2}{c}{15.5\%} & \multicolumn{2}{c}{17.1\%} \\
BS-normalized step time (s) & 65 & 200 & 86 & 240 \\
\bottomrule
\end{tabular}
\caption{Computational cost comparison between OPSD and RL training on 8$\times$H800 GPUs.
OPSD is run for at most 15 steps with effective batch size 1024;
RL is run for over 500 steps with batch size 64.
The BS-normalized step time divides OPSD step time by 16 (= 1024 / 64) to enable a fair per-sample comparison.}
\label{tab:compute_cost}
\end{table}

\end{document}